\RequirePackage{fix-cm}
\documentclass[twocolumn]{svjour3}          % twocolumn
\smartqed  % flush right qed marks, e.g. at end of proof
\usepackage{graphicx}
\usepackage{epsfig}
\usepackage{graphicx}
\usepackage[fleqn]{amsmath}
\usepackage[fleqn]{mathtools}
\usepackage{amssymb}
\usepackage{makecell}
\usepackage{multirow}
\usepackage{url}
\usepackage{algorithm}
\usepackage[noend]{algpseudocode}

\journalname{International Journal of Computer Vision}

\begin{document}

\title{Deep CNN-based Multi-task Learning for Open-Set Recognition}


\author{Poojan Oza         \and
        Vishal M. Patel
}


\institute{Poojan Oza \at
	Whiting School of Engineering \\
	Johns Hopkins University\\
	3400 North Charles Street, Baltimore, MD 21218-2608\\
	\email{poza2@jhu.edu}           
	\and
	Vishal M. Patel \at
	Whiting School of Engineering\\
	Johns Hopkins University\\
	\email{vpatel36@jhu.edu}  
}

\date{Received: date / Accepted: date}

\maketitle

\begin{abstract}
We propose a novel deep convolutional neural network (CNN) based multi-task learning approach for open-set visual recognition.  We combine a classifier network and a decoder network with a shared feature extractor network within a multi-task learning framework. We show that this  approach  results  in  better open-set recogntion accuracy.  In our approach, reconstruction errors from the decoder network are utilized for open-set rejection.  In addition, we model the tail of the reconstruction error distribution from the known classes using the statistical Extreme Value Theory to improve the overall performance.  Experiments on multiple image classification datasets are performed  and it is  shown  that  this  method  can  perform  significantly  better  than many competitive open set recognition algorithms available in the literature.\\The code is available at: github.com/otkupjnoz/mlosr.

\keywords{Open-Set Recognition\and Deep Convolution Neural Network \and Multi-task Learning \and Extreme Value Theory }
\end{abstract}

\section{Introduction}

\begin{figure}[t]
	\begin{center}
		\includegraphics[width=.90\linewidth]{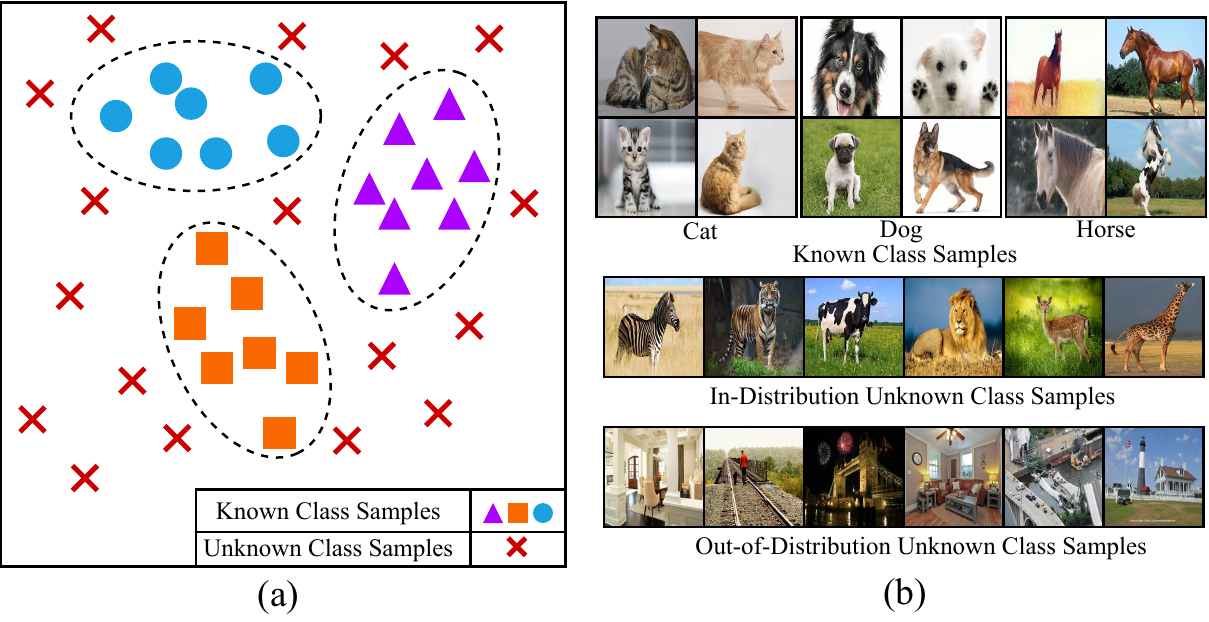}
	\end{center}
 \caption{(a) Typical example of open-set recognition. Here rectangles, triangles, and circles correspond to three different classes known during training and cross samples correspond to unknown class examples that appear only during testing. The goal is to learn a decision boundary such that it discriminates among all three known classes and rejects the unknown class samples. (b) Animal classification example under open-set scenario.   Sample images corresponding to the known classes (i.e. closed set samples) are shown in the first row.  Closed set classes are Cat, Dog, and Horse. The second row shows images corresponding to in-distribution unknown classes i.e., Lion, Zebra, Cow, etc. These animals do not appear during training. The third row shows sample images corresponding to the out-of-distribution unknown classes, i.e., Bridge, Bedroom, House, etc. These images do not contain any animals and are from an entirely different set of categories not seen during training.}
	\label{fig:open_set_figure}
\end{figure}

\begin{figure*}[htp!]
	\begin{center}
		\includegraphics[width=1.0\linewidth]{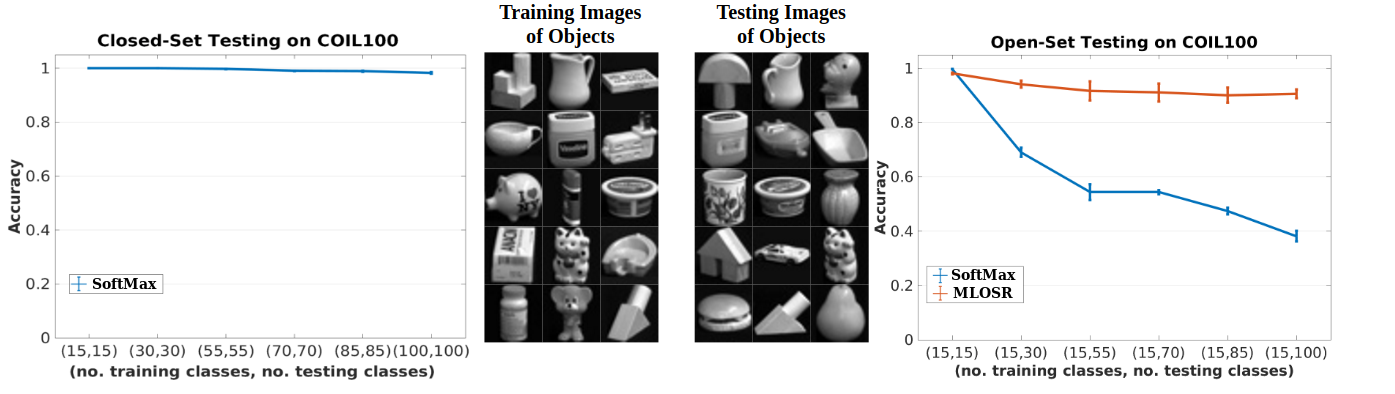}
	\end{center}
	\caption{\textbf{Left:} This graph shows the experimental results of testing a CNN in a closed-set setting on the COIL100 object classification dataset \cite{nene1996columbia}. For the same number of training and testing classes, CNN with SoftMax scores achieves a stable performance of around 100\% accuracy. \textbf{Middle:} Training images of objects shown in the set of images on the left have examples of known categories. For closed-set setting testing images of objects are similar to training images. However, for open-set testing, the images of objects contain a mix of both known and unknown class samples, shown in the set of images on the right. \textbf{Right:} This graph shows the experimental results of testing a CNN in an open-set setting. The number of known classes is kept fix to 15. While, starting with 0 (i.e., closed-set setting) the unknown classes are increased up to 85. It can be seen from the graph that increasing unknown classes during testing significantly affects the performance of a CNN with SoftMax scores, showing the vulnerability of the closed-set model in identifying  unknown classes. On the other hand, the performance of the proposed approach, MLOSR, is much more robust. Jain \emph{et al.} \cite{jain2014multi} perform a  similar experiment for multi-class SVM.}
	\label{fig:coil_open_vs_close}
\end{figure*}

Recent  developments  in  deep  convolutional  neural  networks  (CNNs)  have  shown  impressive  performance on various computer vision tasks such as image classification and object recognition \cite{alexnet12}, \cite{resnet16}, \cite{senet17}. The top-5 error rate of recent image classification methods on ImageNet dataset \cite{imagenet15} has dropped from $\sim$25\% \cite{winner11} to 2.25\% \cite{winner17}. Even though the performance of these systems look very promising, their setting is not realistic. A vast majority of these  algorithms follow a closed-set setting, where the knowledge of all the test classes is assumed to be available during training. However, real-world applications contain many more categories compared to the number of categories present in any of the current datasets, e.g., ImageNet (1000 classes). Hence, when deployed, these systems are highly likely to observe test samples belonging to unknown classes, i.e., classes not observed during training. Because of the closed set assumption, these systems will wrongly recognize a test sample from an unknown class as a sample belonging to one of the known closed set classes.

Open-set recognition was introduced to tackle this problem, extending closed set classification task in a more realistic scenario \cite{scheirer2013toward}. In open-set recognition, an incomplete knowledge of the world is assumed during training, i.e., test samples can be from any of the unknown classes, not observed during training (see Fig.  \ref{fig:open_set_figure}).  The goal of an open-set recognition system is to reject test samples from unknown classes while maintaining the performance on known classes. Since we only have access to the known classes during training phase, it is challenging to identify unknown classes in a closed-set setting. Scheirer \emph{et al.} \cite{scheirer2013toward} proposed a new framework by combining empirical risk minimization with open space risk minimization for open-set recognition problems. Open space risk is defined as the risk of labeling the unknown samples as known. Following this framework, there have been many open-set recognition algorithms proposed over the past few years that try to directly or indirectly minimize open space risk for training open-set recognition models. Jain \emph{et al.} \cite{jain2014multi} showed vulnerability of Support Vector Machine (SVM) based classification (Fig. 1 in the paper \cite{jain2014multi}) in the presence of unknown class test samples and proposed an approach to improve the identification of unknown classes. Many other extensions of the traditional statistical classification approaches for open-set recognition have also been proposed in the literature \cite{scheirer2013toward}, \cite{jain2014multi}, \cite{scheirer2014probability}, \cite{junior2016specialized}, \cite{rudd2018extreme},  \cite{junior2017nearest}, \cite{bendale2015towards}, \cite{zhang2017sparse}.

As mentioned earlier, CNN is a powerful tool to learn discriminative representations for image classification. However, they are fundamentally limited in identifying unknown samples due to their closed-set training (refer to Fig. \ref{fig:coil_open_vs_close} for details). Hence, it is important to make CNN-based image classification algorithms capable of performing open-set recognition. There have been several methods proposed over the years to tackle the presence of unknown classes by extending deep neural networks in open-set settings \cite{bendale2016towards}, \cite{shu2017doc}, \cite{ge2017generative}, \cite{neal2018open}, \cite{yoshihashi2018classification}. Bendale \emph{et al.} \cite{bendale2016towards} proposed to use pre-trained penultimate activations from a neural network and extreme value modeling to update the SoftMax probability values for open-set recognition (referred to as Openmax). Ge \emph{et al.} \cite{ge2017generative} used synthetic unknown classes generated using a Generative Adversarial Network (GAN) \cite{goodfellow2014generative} and trained a neural network to classify those samples as unknown. Shu \emph{et al.} \cite{shu2017doc} proposed a novel loss function by replacing the SoftMax layer with sigmoid activations (referred to as one-vs-rest layer) to train a neural network for open-set recognition. Neal \emph{et al.} \cite{neal2018open} introduced another GAN-based data augmentation approach which generates synthetic unknown class images referred to as counterfactual images for open-set recognition. Yoshihashi \emph{et al.} \cite{yoshihashi2018classification} proposed a novel neural network architecture which involves hierarchical reconstruction blocks and extreme value model for open-set recognition. Though there has been a fair amount of research in developing CNN-based open-set recognition algorithms, the performance of these systems for challenging object recognition datasets is still far from optimal.

In this paper, we present a CNN-based multi-task learning algorithm for open-set recognition. The proposed Multi-task Learning Based Open-Set Recognition (MLOSR) method consists of a shared feature extractor network along with a decoder network and a classifier network for reconstruction and classification, respectively. All these networks are trained in a multi-task learning framework \cite{caruana1997multitask}. We show that such multi-task training yields a better model for open-set recognition by improving the identification of samples from the unknown classes. Additionally, we utilize extreme value theory (EVT) modeling techniques to model the reconstruction error distribution from the network that further enhances the performance. Extensive experiments on multiple image classification datasets show that MLOSR performs better than existing open-set algorithms. In summary, the main contributions of this paper are as follows:

\begin{itemize}
	\item We propose a CNN-based multi-task learning algorithm, called MLOSR, for open-set recognition.
%	\item A proof to show that MLOSR can manage open space risk for deep CNNs (see Appendix).
	\item Extensive experiments on various datasets show that the proposed multi-task training helps to reject out-of-distribution data as well as samples from the in-distribution unknown classes.
\end{itemize}

This paper is organized as follows. Section~\ref{sec:related_work} gives a brief review of open-set recognition and related problems such as out-of-distribution detection, anomaly detection, and EVT. Section~\ref{sec:proposed_approach} introduces the proposed approach and presents training and testing details of the MLOSR algorithm.  Experiments  and  results  are  presented in Section~\ref{sec:experiments_results} and Section~\ref{sec:conclusion} concludes the paper with a brief summary and discussion.

\section{Related Work}\label{sec:related_work}

In this section, we provide some related works on open-set recognition, out-of-distribution detection, EVT and novelty detection.

\subsection{Open-set Recognition} In recent years, a few attempts have been made to create a classifier with rejection option  \cite{bartlett2008classification}, \cite{yuan2010classification}, \cite{da2014learning}. Inspired from these earlier methods, Scheirer \emph{et al.} \cite{scheirer2013toward} formally defined the open-set recognition problem and introduced a framework to train and evaluate such algorithms. Scheirer \emph{et al.} \cite{scheirer2013toward} also introduced a simple slab model-based approach to address this problem. In follow up works by Scheirer \emph{et al.} \cite{scheirer2014probability} and Jain \emph{et al.} \cite{jain2014multi}, both proposed to leverage extreme value models on the SVM decision scores to extend the SVM-based classification in open-set setting. While Jain \emph{et al.} \cite{jain2014multi} utilized the decision scores only from One-vs-All multi-class SVM, Scheirer \emph{et al.} \cite{scheirer2014probability} combined the scores from multi-class SVM with class-specific one-class RBF-SVMs to get a better open-set model. Junior \emph{et al.} \cite{junior2017nearest} proposed a nearest neighbor-based classification approach based on the similarity scores calculated using the ratio of distances between the nearest neighbors, and identified any test sample as unknown having low similarity. Zhang and Patel  \cite{zhang2017sparse} proposed another approach by extending the sparse representation-based classification (SRC) to the open-set setting. They also discovered that the residual errors from SRC contain some discriminative information to identify known and unknown classes. These residual errors are modeled using EVT as match and non-match to identify unknown test samples by hypothesis testing. 

Following these extensions of traditional classification algorithms for open-set recognition, Bendale \emph{et al.} \cite{bendale2016towards} became the first work to address the open-set recognition problem for deep neural networks. Since, thresholding on SoftMax probability does not yield a good model for open-set recognition \cite{bendale2016towards} (also shown in Fig. \ref{fig:coil_open_vs_close}), an alternative solution was proposed for adapting deep neural network to open-set settings. Bendale \emph{et al.} \cite{bendale2016towards} utilized the activation vectors from a penultimate layer of a pre-trained deep neural network. Modeling distance of these activation vectors from the mean of each class with EVT an updated penultimate vector is generated (referred to as OpenMax). This updated vector yields a better model for identifying unknown class test samples. Ge \emph{et al.} \cite{ge2017generative} introduced G-OpenMax algorithm which combines OpenMax with data augmentation using GANs. Ge \emph{et al.} generated unknown samples from the known class data using GANs and later used them for training a CNN along with known classes. This data augmentation technique was shown to improve the unknown class identification. In another approach, Shu \emph{et al.} \cite{shu2017doc} argued that OpenMax inherently considers that hard to classify samples are more likely to be from the unknown classes and proposed a $K$-sigmoid activation-based method, to overcome that issue. The $K$ sigmoid activation method replaces the SoftMax layer to train the network with a novel loss function. Neal \emph{et al.} introduced another GAN-based data augmentation method. Instead of considering the misclassified samples generated using GAN as unknown classes like G-OpenMax, Neal \emph{et al.} proposed a method to search for such examples, referred to as counterfactual-images. These counterfactual samples are later augmented with the original dataset as unknown class samples and are utilized to fine-tune the classification network. This technique was shown to be a better data augmentation approach than G-OpenMax for open-set recognition. Recently, Yoshihashi \emph{et al.} \cite{yoshihashi2018classification} proposed a novel neural network architecture for open-set recognition which consists of hierarchical reconstruction modules combined with extreme value modeling. To the best of our knowledge, it is the best performing open-set algorithm in the literature.

\subsection{Out-of-Distribution Detection} Recently, some concerns have been raised regarding the safety of AI systems \cite{amodei2016concrete}. These include Adversarial Attacks \cite{szegedy2013intriguing}, \cite{yuan2017adversarial}, Distribution Shifts \cite{hendrycks2016baseline}, \cite{lee2017training} etc. One of the concerns regarding the distribution shift is the detection of out-of-distribution (OOD) examples. Hendrycks \emph{et al.} first introduced the problem of OOD in \cite{hendrycks2016baseline} and defined it as the detection of samples that do not belong in the training set but can appear during testing.  Several approaches have been proposed in the literature that address the OOD problem \cite{hendrycks2016baseline}, \cite{liang2017enhancing}, \cite{hendrycks2018deep}, \cite{lee2017training}, \cite{liu2018open}. The experimental setup used for evaluating OOD usually includes two datasets, a clean set with finite categories available during training,  and testing as well as OOD test set containing samples from a completely different distribution. For example, classification on CIFAR10 \cite{krizhevsky2014cifar} will have OOD examples from LSUN \cite{yu2015lsun}. Both OOD and open-set problems are studied separately even though the OOD problem setting resembles that of open-set recognition.  Furthermore, the evaluation protocol followed by both OOD and open-set recognition problems is very similar. Hence, we will also explore the capability of the proposed open-set recognition method in detecting out-of-distribution samples.

\subsection{Outlier-Novelty-Anomaly Detection} The problems such as discovering outliers \cite{xia2015learning}, \cite{you2017provable}, \cite{niu2018learning}, identifying novel classes \cite{abati2018and}, \cite{perera2018learning}  and detecting anomalies \cite{golan2018deep}, \cite{sabokrou2018adversarially}, \cite{chalapathy2017robust} also have some overlap with open-set recognition. Though all of these problems involve identifying abnormality/novelty, the problem setting differs from the open-set problem. Though anomaly/novelty detection problems do not have access to abnormal/novel classes during training phase, several works assume the availability of abnormal classes \cite{hendrycks2018deep} during training and they are mainly limited to one class recognition problems. Outlier detection allows access to outlier data during the training phase. On the other hand, open-set recognition problems do not have access to unknown class data and also deals with multi-class classification problem and hence is more challenging than outlier, novelty or anomaly detection.

\subsection{Extreme Value Theory (EVT)} Extreme Value Theory has proven to be useful in many vision applications \cite{shi2008modeling}, \cite{gibert2017deep}, \cite{scheirer2017extreme}, \cite{scheirer2011meta}, \cite{fragoso2013evsac} including open-set recognition \cite{scheirer2014probability}, \cite{rudd2018extreme}, \cite{yoshihashi2018classification}, \cite{bendale2016towards}, \cite{zhang2017sparse}. This popularity is attributed by the fact that extreme value modeling of decision scores yields better performance than directly utilizing the raw score values \cite{scheirer2011meta}, \cite{scheirer2017extreme}. Extreme value modeling has became one of the most popular approaches for post recognition score analysis to improve the performance of open-set recognition. Inspired by these methods, the proposed approach also utilizes EVT to obtain better recognition scores.

%%%%%%%%%%%%%%%%%%%%%%%%%%%%%%%%%%%%%%%%%%%%%%%%%%%%%%%%%%%%%%%%%%%%%%%%%%%%%%%%%%%%%%%%%%%\hfill mds
%%%%%%%%%%%%%%%%%%%%%%%%%%%%%%%%%%%%%%%%%%%%%%%%%%%%%%%%%%%%%%%%%%%%%%%%%%%%%%%%%%%%%%%%%%% 
%%%%%%%%%%%%%%%%%%%%%%%%%%%%%%%%%%%%%%%%%%%%%%%%%%%%%%%%%%%%%%%%%%%%%%%%%%%%%%%%%%%%%%%%%%%\hfill December XX, 2018

\section{Proposed Approach}\label{sec:proposed_approach}

%This section starts with an overview of the basic block diagram of system modules, followed by a discussion regarding the training strategy used to train neural network modules. In the end, we describe the extreme value model used for modeling reconstruction errors with a brief discussion of EVT.
%\subsection{Block Diagram}\label{subsec:block_diagram}

\begin{figure*}[t!]
	\begin{center}
		\includegraphics[width=0.75\linewidth]{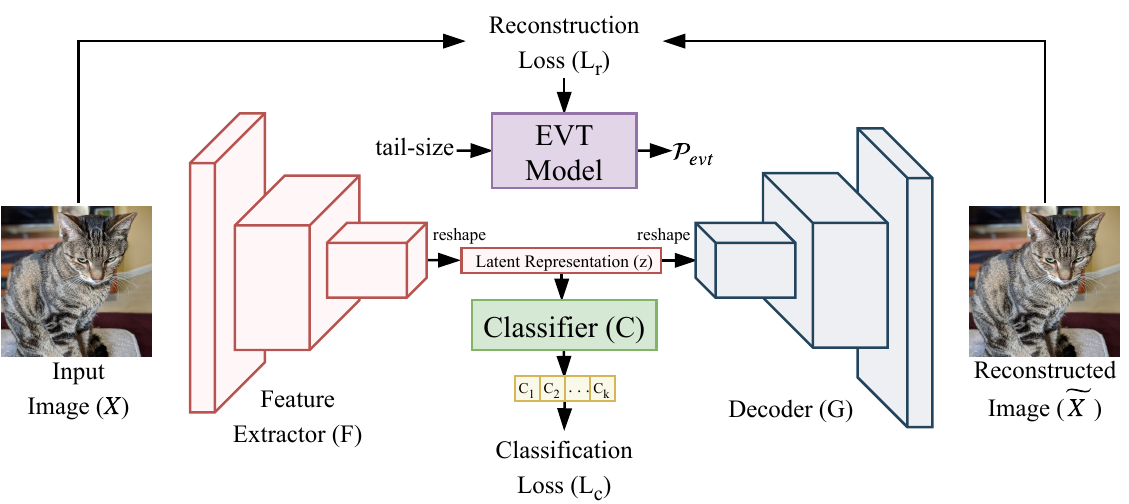}
	\end{center}
	\caption{An overview of the proposed MLOSR algorithm. The feature extractor network ($\mathcal{F}$) takes any input image ($X$) and produces a latent representation $(z)$. This latent representation is used by the classifier ($\mathcal{C}$) and the decoder ($\mathcal{G}$) to predict class label and to reconstruct the input ($\tilde{X}$), respectively. These networks are trained to perform both classification and reconstruction task in a multi-task framework. The tail of the reconstruction error distribution is modeled using EVT. The classification score and the EVT probability of reconstruction error is used during testing to perform open set recognition.}
	\label{fig:block_diagram}
\end{figure*}

In this section, we describe the proposed approach in detail.  The proposed system architecture is composed of four modules:\\ 
\noindent \textbf{1.} Feature Extractor (or Encoder) ($\mathcal{F}$),\\
\noindent \textbf{2.} Decoder ($\mathcal{G}$),\\ 
\noindent \textbf{3.} Classifier ($\mathcal{C}$), and\\
\noindent \textbf{4.} Extreme Value Model ($\mathcal{P}_{evt}$).\\ Fig.~\ref{fig:block_diagram} shows these modules of the proposed system. The feature extractor (or encoder) network ($\mathcal{F}$) is modeled by a CNN architecture which maps an input image onto a latent space representation. The decoder ($\mathcal{G}$), modeled by another CNN and a classifier ($\mathcal{C}$), modeled by a fully-connected neural network, take this latent representation as input and produce a reconstructed image and its label as the outputs, respectively. Both the decoder network and the classifier network share the feature extractor module. After the models $\mathcal{F}$, $\mathcal{G}$ and $\mathcal{C}$ are trained, the reconstruction errors are modeled using EVT. In the following sections, we present the training procedure to learn the parameters ($\Theta_f$, $\Theta_g$, $\Theta_c$) and discuss the recognition score analysis using EVT.

\subsection{Training Procedure}\label{subsec:training_procedure}
The feature extractor network can be represented by a function, $\mathcal{F}: \mathcal{X} \rightarrow \mathcal{Z}$.  Similarly, let the decoder and the classifier networks be represented by functions $\mathcal{G}: \mathcal{Z} \rightarrow \mathcal{X}$ and $\mathcal{C}: \mathcal{Z} \rightarrow \mathcal{Y}$, respectively.  Here, $\mathcal{Z}$ is the space of latent representations, $\mathcal{X}$ is the space of all images, and $\mathcal{Y}$ is the space of all $K$ possible image labels. Let $\Theta_f$, $\Theta_g$ and $\Theta_c$ be the parameters of $\mathcal{F}$, $\mathcal{G}$ and $\mathcal{C}$, respectively.  The classification loss, denoted by $\mathcal{L}_c$, penalizes the network for misclassifying known class samples. The reconstruction loss, denoted by $\mathcal{L}_r$, penalizes the network for generating images away from the known class samples. Let  $X_i \in \mathcal{X}$ and $y_i \in \mathcal{Y}$ be a sample from any of the known classes and it's corresponding label. Let  $\tilde{X}_i = \mathcal{G}(\mathcal{F}(X_i))$ be the reconstructed input from the encoder-decoder pipeline, $\mathcal{F}$+$\mathcal{G}$. Also, $\tilde{y}_i = \mathcal{C}(\mathcal{F}(X_i))$ is the predicted class probability vector by the encoder-classifier pipeline, $\mathcal{F}$+$\mathcal{C}$. The loss function $\mathcal{L}_r$ depends on the parameters $\Theta_f$, $\Theta_g$ which are associated with the networks $\mathcal{F}$ and $\mathcal{G}$, respectively. Similarly, the loss function $\mathcal{L}_c$ depends on the parameters $\Theta_f$, $\Theta_c$ which are associated with the networks $\mathcal{F}$ and $\mathcal{C}$, respectively. We can formulate the losses for the input images with batch size of $N$ as follows
\begin{equation}
\begin{aligned}
\mathcal{L}_c(\{\Theta_f, \Theta_c\}) \ = \ \frac{1}{N} \sum_{i=1}^{N} \ell_c(\mathbb{I}_{y_i}, \tilde{y_i};\{\Theta_f, \Theta_c\}), 
\end{aligned}
\label{eq:lc}
\end{equation}
\begin{equation}
\begin{aligned}
\mathcal{L}_r(\{\Theta_f, \Theta_g\}) \ = \ \frac{1}{N} \sum_{i=1}^{N} \ell_r(X_i, \tilde{X_i};\{\Theta_f, \Theta_g\}), 
\end{aligned}
\label{eq:lr}
\end{equation}
where, $\mathbb{I}_{y_i}$ is a one-hot vector for label $y_i$. Also, $\ell_r$ and $\ell_c$ can be implemented using any valid classification and reconstruction loss functions, respectively. For this method we consider cross-entropy for the classification loss ($\ell_c$) and $L^1$-norm of the vectorized images for the reconstruction loss ($\ell_r$). $\ell_r$ and $\ell_c$ are defined as follows,
\begin{equation}
\begin{aligned}
\ell_c(y_i, \tilde{y_i}) \ = \ -\sum_{j=1}^{K} \ \mathbb{I}_{y_i}(j) \  \text{log} [ \tilde{y_i}(j) ], 
\end{aligned}
\end{equation}

\begin{equation}
\begin{aligned}
\ell_r(X_i, \tilde{X_i}) \ = \ \| \ X_i - \tilde{X_i} \ \|_1,
\end{aligned}
\end{equation}
The final loss to train the overall network is as follows
\begin{equation}
\begin{aligned}
& \underset{\{\Theta_f, \Theta_g, \Theta_c\}}{\text{min}}
& &  \lambda_c \mathcal{L}_c(\{\Theta_f, \Theta_c\}) + \lambda_r \mathcal{L}_r(\{\Theta_f, \Theta_g\}).
\end{aligned}
\label{eq:finalobj}
\end{equation}
Here, $\lambda_r$ and $\lambda_c$ are two constants and $K$ is the total number of classes known during training. After training, the learned parameters $\tilde{\Theta}_f$, $\tilde{\Theta}_c$, $\tilde{\Theta}_g$ will yield an open-set recognition model.

\subsection{Extreme Value Model}\label{subsec:extreme_value_model}

\begin{figure}[b!]
	\centering
	\includegraphics[width=0.53\textwidth]{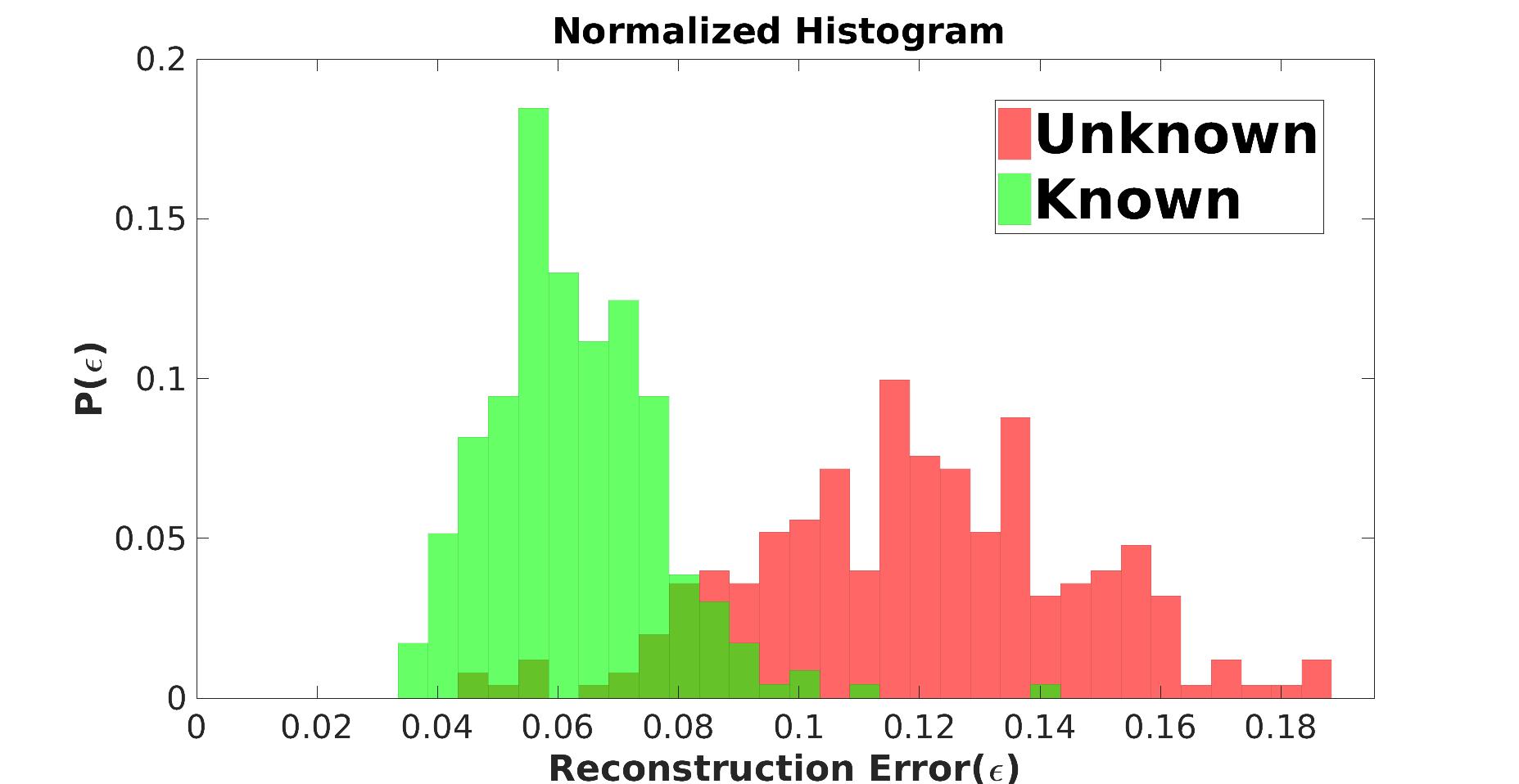}
	\caption{Reconstruction error distributions for known and unknown classes, computed using the COIL-100 dataset.}
	\label{fig:histogram}
\end{figure}

As discussed in Section \ref{sec:related_work}, EVT is useful in many vision applications. Some open-set recognition algorithms \cite{scheirer2014probability}, \cite{rudd2018extreme}, \cite{yoshihashi2018classification}, \cite{bendale2016towards}, \cite{zhang2017sparse} also utilize this tool to model the tail part of the distribution corresponding to the recognition scores. Histograms corresponding to the reconstruction errors from both known (shown in green) and unknown (shown in red) class samples computed using the COIL-00 dataset are shown in Fig.~\ref{fig:histogram}.  As can be seen from this figure that the reconstruction errors contain some information to discriminate between unknown and known classes. Since, during training we do not have access to the samples from unknown classes, the region of optimal decision threshold must lie somewhere within the set of extremes values from the known class reconstruction errors (the overlapped region). With this observation, we also use EVT to model the tail part of the reconstruction error distribution to achieve a better estimate of the tail data. 

\begin{algorithm}[t!]
	\caption{Pseudocode for MLOSR Training}\label{algo:mlosr_training}
	\begin{algorithmic}[1]
		\Require Network models $\mathcal{F}$, $\mathcal{C}$, $\mathcal{G}$
		\Require Initial respective parameters $\Theta_f$, $\Theta_c$, $\Theta_g$
		\Require Labeled known data, $\{(X_i, y_i)\}_{i=1}^{i=N_s}$
		\Require Hyper-parameters : $N$, $\eta$, $\lambda_c$, $\lambda_r$
		\State Latent space representation, $z = \mathcal{F}(X)$
		\State Prediction probabilities,  $p_{y} = \mathcal{C}(z)$
		\State predict known label,  $y_{pred} = argmax(p_y)$
		\State \textbf{while} not done \textbf{do}
		\State $ \ \ \ \ \ $ \textbf{for} each batch with size $N$ \textbf{do}
		\State $ \ \ \ \ \ \ \ \ \ \ $ \textbf{for} i = $1$ to $N$ \textbf{do}		
		\State $ \ \ \ \ \ \ \ \ \ \ \ \ \ \ \ z_{i} \ = \mathcal{F}(X_i)$
		\State $ \ \ \ \ \ \ \ \ \ \ \ \ \ \ \ \tilde{y}_i \ = \mathcal{C}(z_{i})$
		\State $ \ \ \ \ \ \ \ \ \ \ \ \ \ \ \ \tilde{X}_i = \mathcal{G}(z_{i})$
		\State $ \ \ \ \ \ \ \ \ \ \ $ \textbf{end for}
		\State $ \ \ \ \ \ \ \ \ \ \ $ Calculate $\mathcal{L}_c$, $\mathcal{L}_r$ using Eq. \ref{eq:lc}, and \ref{eq:lr}
		\State $ \ \ \ \ \ \ \ \ \ \ \ \mathcal{L}_t = \lambda_c \mathcal{L}_c + \lambda_r \mathcal{L}_r$
		\State $ \ \ \ \ \ \ \ \ \ \ $ Update $\Theta_c$ : $\Theta_c \leftarrow \Theta_c - \eta \nabla_{\Theta_c}\mathcal{L}_t$  
		\State $ \ \ \ \ \ \ \ \ \ \ $ Update $\Theta_g$ : $\Theta_g \leftarrow \Theta_g - \eta \nabla_{\Theta_g}\mathcal{L}_t$
		\State $ \ \ \ \ \ \ \ \ \ \ $ Update $\Theta_f$ : $\Theta_f \leftarrow \Theta_f - \eta \nabla_{\Theta_f}\mathcal{L}_t$
		\State $ \ \ \ \ \ $ \textbf{end for}
		\State \textbf{end while}
		\State \textbf{Output:} Learned parameters $\tilde{\Theta}_f$, $\tilde{\Theta}_c$, $\tilde{\Theta}_g$
	\end{algorithmic}
\end{algorithm}

There are two widely used theorems to model statistical extremes, namely, Fisher-Tippett-Gnedenko theorem (FTG) \cite{scheirer2017extreme} and Picklands-Balkema-deHaan formulation or Generalized Extreme Value theorem (GEV) \cite{pickands1975statistical}, \cite{balkema1974residual}. Some works follow FTG \cite{scheirer2014probability}, \cite{bendale2016towards}, \cite{yoshihashi2018classification} while others follow GEV \cite{zhang2017sparse}. In this method, we consider extreme value formulation by Picklands-Balkema-deHaan or GEV. It states that for large enough threshold $w$, for a large class of distributions denoted as $V$, with $\{V_1,..,V_n \}$, $n$ IID samples, the following equation is well approximated by a Generalized Pareto Distribution (GPD), denoted as $\mathcal{P}_{evt}$,
%\vskip -10.0pt
\begin{align*}
Pr(v - w \leq v | v > w) &= \frac{F_V(w+v)-F_V(v)}{1-F_V(w)} \\ &= \mathcal{P}_{evt}(v;\zeta, \mu)
\label{eq:gev}
\end{align*}
where,
\begin{equation*}
\mathcal{P}_{evt}(v;\zeta, \mu)=
\begin{cases}
1 - (1+ \frac{\zeta \cdot v}{\mu})^{\frac{1}{\zeta}}, \ \text{if} \ \zeta \neq 0,\\
1 - e^{\frac{v}{\mu}} \ \ \ \ \ \ \ \ \ \ \ \ \ , \ \text{if} \ \zeta = 0,\\
\end{cases}
\end{equation*}
such that $-\infty < \zeta < +\infty$, $0 < \mu < +\infty$, $v > 0$ and $\zeta w > -\mu$. $\mathcal{P}_{evt}(\cdot)$ is CDF of GPD and hence, $\mathcal{P}_{evt}(r)$ denotes the probability that extremes of distribution $V$ is less than some value $r > w$. This probability score will be useful in making decision about whether a given test sample is from known classes or not. The parameters $\zeta$ and $\mu$ can be estimated from the given tail data, using maximum likelihood estimation procedure, provided by Grimshaw \emph{et al.} \cite{grimshaw1993computing}. Here, there are two user defined parameters - the tail size and the threshold value to make the decision on known/unknown classes (more information is provided in Section \ref{sec:experiments_results}).  The MLOSR training and testing procedures are summarized in Algorithms \ref{algo:mlosr_training} and \ref{algo:mlosr_testing}, respectively. 

\begin{algorithm}[t!]
	\caption{Pseudocode for MLOSR Testing}\label{algo:mlosr_testing}
	\begin{algorithmic}[1]
		\Require Trained network models $\mathcal{F}$, $\mathcal{C}$, $\mathcal{G}$
		\Require Learned parameters $\tilde{\Theta}_f$, $\tilde{\Theta}_c$, $\tilde{\Theta}_g$	
		\Require EVT model $\mathcal{P}_{evt}$, threshold $\tau$
		\Require Test image $X$, either from known or unknown data
		\State Latent space representation, $z = \mathcal{F}(X)$
		\State Prediction probabilities,  $p_{y} = \mathcal{C}(z)$
		\State predict known label,  $y_{pred} = argmax(p_y)$
		\State Reconstructed Image,  $\tilde{X} = \mathcal{G}(z)$
		\State Reconstruction Error, $\text{r} = ||X-\tilde{X}||_1$
		\State \textbf{if} $ \ \ \mathcal{P}_{evt}(r) \ < \ \tau \ $ \textbf{do}
		\State $ \ \ \ \ $ predict $X$ as Known, with label $y_{pred}$ %known with label $y_{pred}$
		\State \textbf{else do}
		\State $ \ \ \ \ $ predict $X$ as Unknown
		\State \textbf{end if}
	\end{algorithmic}
\end{algorithm}

%%%%%%%%%%%%%%%%%%%%%%%%%%%%%%%%%%%%%%%%%%%%%%%%%%%%%%%%%%%%%%%%%%%%%%%%%%%%%%%%%%%%%%%%%%%\hfill mds
%%%%%%%%%%%%%%%%%%%%%%%%%%%%%%%%%%%%%%%%%%%%%%%%%%%%%%%%%%%%%%%%%%%%%%%%%%%%%%%%%%%%%%%%%%% 
%%%%%%%%%%%%%%%%%%%%%%%%%%%%%%%%%%%%%%%%%%%%%%%%%%%%%%%%%%%%%%%%%%%%%%%%%%%%%%%%%%%%%%%%%%%\hfill December XX, 2018

\section{Experiments and Results}\label{sec:experiments_results}
In this section, we demonstrate the effectiveness of the proposed MLOSR approach by conducting various experiments on the  COIL-100 \cite{nene1996columbia}, MNIST \cite{lecun2010mnist}, SVHN \cite{netzer2011reading}, CIFAR10 \cite{krizhevsky2009learning} and TinyImageNet \cite{le2015tiny} datasets.  In particular, we first present analysis of the proposed approach on the COIL-100 dataset.  Then, we compare the performance of MLOSR with recent state-of-the-art open-set recognition algorithms on four image classification datasets (MNIST, SVHN, CFAR10, TinyImageNet). In these experiments, unknown classes are sampled from within the dataset by dividing the total number of classes into known and unknown categories. In the last set of experiments, we test the ability of MLOSR to detect out-of-distribution unknown samples. In this experiment, the in-distribution samples are from CIFAR10 and out-of-distribution samples are from ImageNet and LSUN, as provided by \cite{liang2017enhancing}. 

The networks are trained using the Adam optimizer \cite{kingma2015adam} with the learning rate ($\eta$) of 0.0003 and batch size ($N$) of 64. We stop the training when $\mathcal{L}_r$ loss becomes sufficiently small. Both weights $\lambda_r$ and $\lambda_c$ for reconstruction and classification loss are set equal to 0.5. For EVT modeling, we keep the tail-size of 20 for all experiments. The decision threshold during testing is set equal to 0.5, i.e., identify any sample with reconstruction error $r$ and probability $\mathcal{P}_{evt}(r)$ less than 0.5 as unknown.

\subsection{Experiment I: Analysis of MLOSR}\label{subsec:preliminary_analysis}
In this experiment, we perform the quantitative and the qualitative analysis to give insights into the proposed MLOSR algorithm. For  quantitative analysis, we measure the performance gain contributed by each module of the overall algorithm. The qualitative study provides visual examples of the reconstructed known and unknown test samples.

\subsubsection{Quantitative Analysis}\label{subsubsec:quantitative_analysis}

\begin{figure}[b!]
	\centering
	\includegraphics[width=0.525\textwidth]{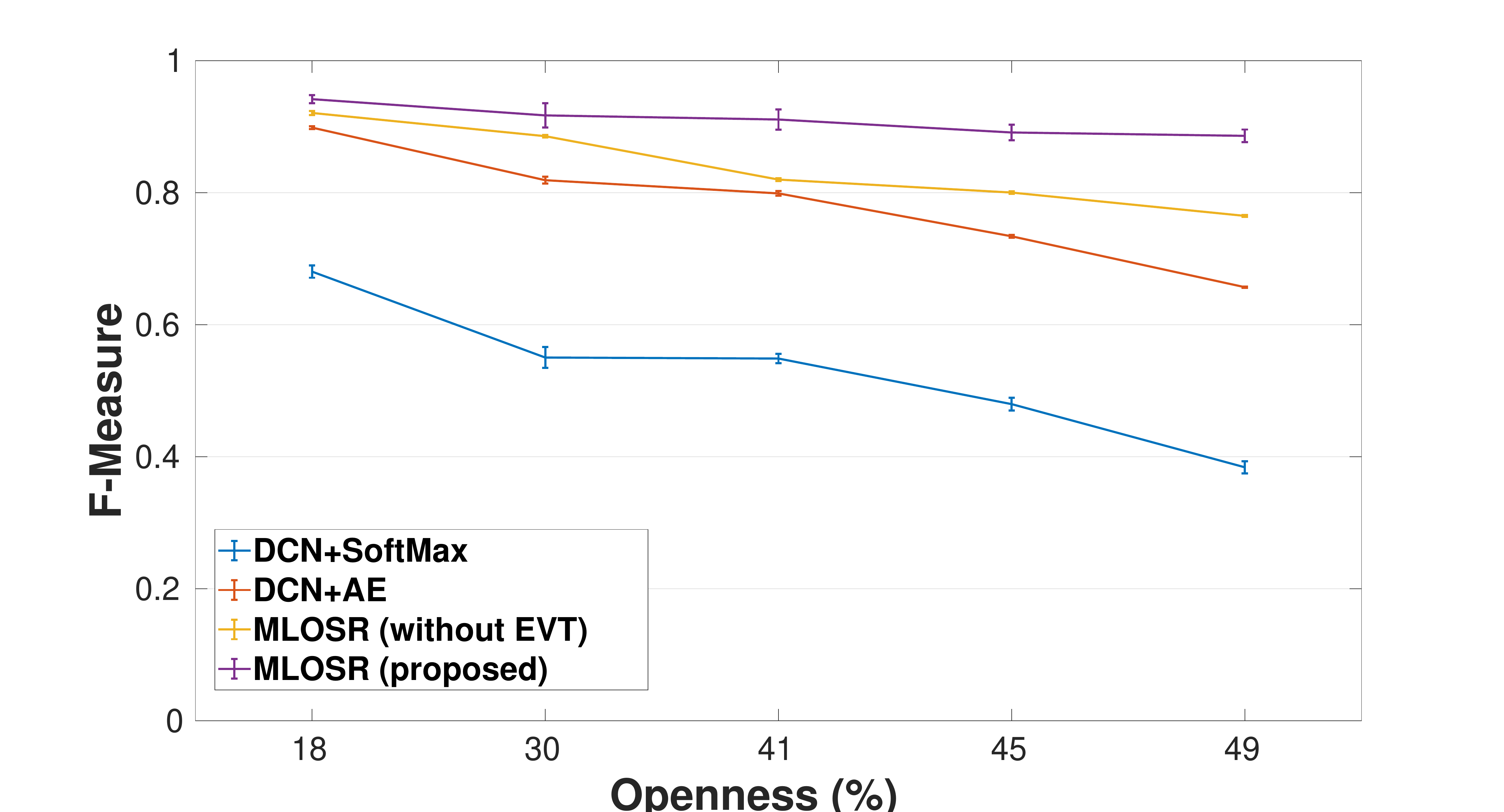}
	\caption{Quantitative Analysis: Ablation study of MLOSR on the COIL-100 dataset.}
	\label{fig:ablation_coil100}
\end{figure}

We perform ablation experiments showing the contribution from each modules of MLOSR on the COIL-100 dataset. The COIL-100 dataset has 100 different object categories with varying pose. Out of 100 classes randomly sampled 15  classes are used as known classes and the remaining are used as unknown. The dataset contains colored images of size $128 \times 128 \times 3$. For this experiment, each image is converted into gray-scale, resized to $64 \times 64$ and intensity values are normalized between $[-1, 1]$. The network architectures used for this experiments are as follows, \\
\textbf{Encoder:}Conv(32)-ReLU-Conv(64)-ReLU-Conv(128)-FC(512). \\ \textbf{Decoder:}FC(8192)-ConvTran(64)-ReLU-ConvTran(32)-ReLU-ConvTran(1)-Tanh. \\ \textbf{Classifier:}FC(512)-FC(15).\\ Here, Conv(L) and ConvTran(L) denote a convolution and transposed-convolution layer with L filters of size $3 \times 3$ and stride 2 respectively, ReLU and Tanh are activation units and FC(L) denote fully connected layer with L neurons.

For ablation analysis, the performance is measured using F-measure (or F1 score) against varying Openness \cite{scheirer2013toward} of the problem. Openness is defined as follows
\begin{equation}\label{eq:openness}
\begin{aligned}
Openness = 1 \ - \ \sqrt{\frac{2\times N_{train}}{N_{test}+N_{target}}},
\end{aligned}
\end{equation}
Here, $N_{target}$ is the number of target classes, $N_{train}$ is the number of train classes and $N_{test}$ is the number of test classes \cite{scheirer2013toward}.  For this experiment openness is varied by keeping 15 classes as known and changing the number of unknown classes from 15 to 85.  This corresponds to the change in openness from 18\% to 49\%. The performance and the corresponding errors for each openness are calculated as average and standard deviation of five randomized trials.

We consider the following methods as baselines for comparison:\\
\textbf{1. DCN+SoftMax:} Encoder and classifier networks are trained using the classification loss $\mathcal{L}_c$ and the SoftMax scores are used for closed-set classification. A test sample is identified as unknown if the leading SoftMax score (between $[0, 1]$) is less than 0.5. This baseline is a traditional closed-set model with a threshold over the SoftMax scores.\\
\textbf{2. DCN+AE:} Encoder and classifier networks are trained using $\mathcal{L}_c$ and SoftMax scores are used for classification. However, to identify any test sample as unknown, an auto-encoder is used, with encoder-decoder architectures as described above. This encoder-decoder pipeline is trained with a reconstruction loss of $\mathcal{L}_r$. A test sample is identified as unknown when the reconstruction error is more than 50\% of the maximum reconstruction error observed on training samples. Another difference to note is that the encoder network is not shared across encoder-classifier and encoder-decoder pipelines. Instead, these pipelines are trained separately with two different encoder networks having the same architecture. This baseline shows the use of reconstruction error as a score compared to SoftMax scores for identifying unknown classes. It also provides a baseline to compare with multi-task training having a shared encoder network.\\
\textbf{3. MLOSR (without EVT):} Encoder-classifier and encoder-decoder pipelines are trained with a shared encoder network using $\mathcal{L}_c$ and $\mathcal{L}_r$ loss function in a multi-task fashion. SoftMax scores and reconstruction errors are utilized for closed-set classification and identifying unknown classes, respectively. Method to identify any test sample as unknown is similar to the method mentioned in the previous baseline. This method provides a baseline to compare the performance with and without extreme value modeling and shows benefits of multi-task training.\\
\textbf{4. MLOSR (Proposed):} This is the method proposed in this paper, where after multi-task training of encoder, classifier and decoder networks, EVT models the tail part of the reconstruction errors from known classes as described in Section \ref{subsec:extreme_value_model}. A test sample is identified as unknown when the reconstruction error has less than 0.5 probability of coming from a known class. Extreme value model ($\mathcal{P}_{evt}$) provides the probability score.

From Fig.~\ref{fig:ablation_coil100} it is clear that \textbf{DCN+SoftMax} is not an optimal model for open-set recognition and has the worst performance among all the baselines. \textbf{DCN+AE} shows that utilizing reconstruction errors from an encoder-decoder trained on known classes elevates the performance of the open-set recognition model. It shows that the reconstruction errors are better than SoftMax scores for identifying unknown classes. Furthermore \textbf{MLOSR (without EVT)} shows that \textbf{DCN+AE} performance can be further improved by utilizing a multi-task training strategy with shared encoder, and improves the open-set performance of the model even further. Finally, the \textbf{MLOSR} utilizes the extreme value model on the known class to better model the tail part of the reconstruction error distribution. This in turn gives improvements over \textbf{MLOSR (without EVT)} under varying openness.

\subsubsection{Qualitative Analysis}\label{subsubsec:qualitative_analysis}

\begin{figure}[htp!]
	\centering
	\includegraphics[width=.46\textwidth]{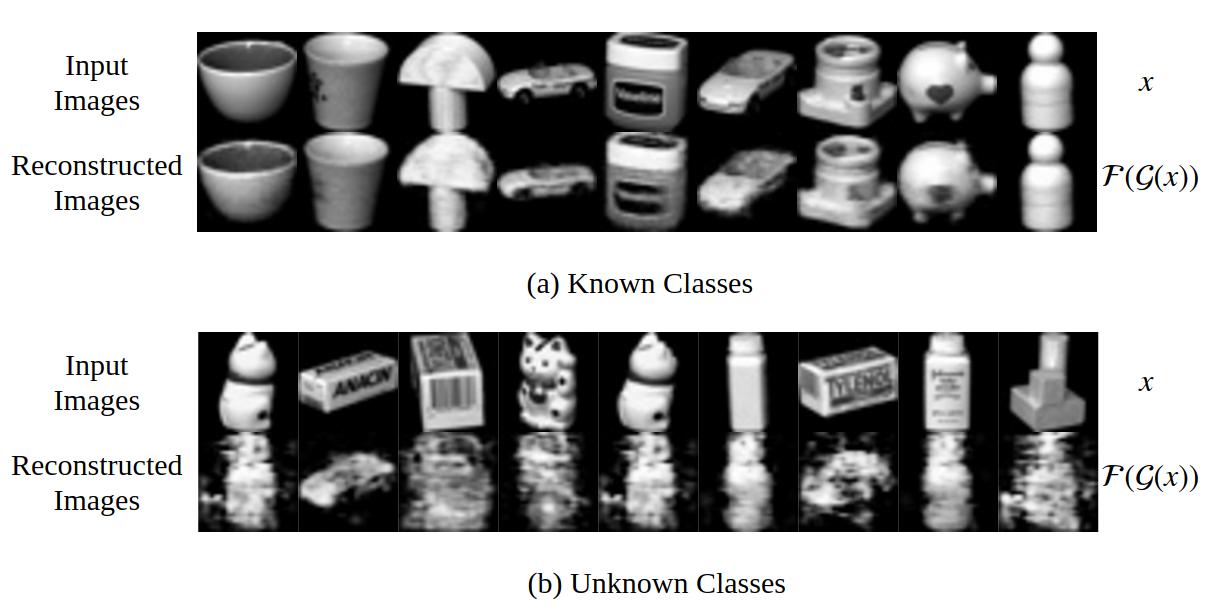}
	\caption{Qualitative Analysis: (a) Known and (b) Unknown class samples from the COIL100 dataset with the corresponding images reconstructed by MLOSR.}
	\label{fig:qualitative_coil100}
\end{figure}

Fig.~\ref{fig:qualitative_coil100} shows the qualitative analysis of the MLOSR algorithm in cases where the test input is from known and unknown classes. Models trained with MLOSR produce output that correctly reconstructs the test input if they are from the known classes, resulting in low reconstruction errors. On the other hand, for the test samples from unknown classes, MLOSR produces distorted outputs resulting in high reconstruction errors.

\subsection{Experiment II: Open-set Recognition}\label{subsec:openset_experiments}
For the open-set recognition experiments, we use the testing protocol followed in \cite{neal2018open}. For the encoder, decoder and classifier networks, the architectures are also the same as \cite{neal2018open}. For all other methods compared, the same architecture is also followed for this experiment.  All images are resized to $32 \times 32$ and intensity values are normalized between $[-1, 1]$ for each color channel. The following experimental protocols are followed:

\noindent \textbf{\textit{(i)} MNIST:} MNIST \cite{lecun2010mnist} has total 10 digit classes having images of handwritten digits. Out of which the number of randomly chosen known and unknown classes used are 6 and 4, respectively. This choice results in the openness of 13.39\%.\\
\textbf{\textit{(ii)} SVHN:} SVHN \cite{netzer2011reading} is a digit dataset, where the images of each digit is cropped from house number image data collected from the google street view images. SVHN has a total of 10 digit classes,  Similar to MNIST, for SVHN randomly chosen  6 known and 4 unknown classes are used with the openness of 13.39\%.\\
\textbf{\textit{(iii)} CIFAR10:} CIFAR10 \cite{krizhevsky2014cifar} consists of 10 object categories. Out of which, we randomly choose 6 known and 4 unknown classes which results in the openness of 13.39\%.\\
\textbf{\textit{(iv)} CIFAR+10:} CIFAR+10 uses 4 classes from CIFAR10 that are non animal categories and 10 classes from CIFAR100 \cite{krizhevsky2014cifar} are randomly sampled from the animal categories as known and unknown classes, respectively. This results in the openness of 33.33\%.\\
\textbf{\textit{(v)} CIFAR+50:} Similar to CIFAR+10, 4 non-animal CIFAR10 categories as sampled as known classes and 50 animal categories from CIFAR100 as sampled as unknown, resulting in the openness of 62.86\%.\\
\textbf{\textit{(vi)} TinyImageNet:} TinyImageNet \cite{le2015tiny} dataset is derived from ImageNet \cite{imagenet15} by reducing the number of classes and image sizes. It has a total of 200 categories with 500 images per category for training and 50 for testing. From 200 categories 20 known and 180 unknown classes are randomly sampled, resulting in openness of 57.35\%.

\begin{table*}[htp!]
	\centering
	\begin{tabular}{|c|c|c|c|c|c|c|}
		\hline
		\textbf{Method}   & \textbf{MNIST} & \textbf{SVHN} & \textbf{CIFAR10} & \textbf{CIFAR+10} & \textbf{CIFAR+50} & \textbf{TinyImageNet} \\ \hline
		SoftMax           & 0.978 & 0.886 & 0.677 & 0.816 & 0.805 & 0.577 \\ \hline
		OpenMax \cite{bendale2016towards} (CVPR'16)   & 0.981 & 0.894 & 0.695 & 0.817 & 0.796 & 0.576 \\ \hline
		G-OpenMax \cite{ge2017generative} (BMVC'17) & 0.984 & 0.896 & 0.675 & 0.827 & 0.819 & 0.580 \\ \hline
		OSRCI \cite{neal2018open} (ECCV'18)     & 0.988 & 0.910 & \textit{0.699} & \textit{0.838} & \textit{0.827} & 0.586 \\ \hline
		CROSR \cite{yoshihashi2018classification} (CVPR'19) & \textbf{0.998} & \textbf{0.955} & ------ & ------ & ------ & \textit{0.670} \\ \hline
		MLOSR  & \textit{0.989} & \textit{0.921} & \textbf{0.845} & \textbf{0.895} & \textbf{0.877} & \textbf{0.718} \\ \hline
	\end{tabular}
	\caption{Comparison of MLOSR with the most recent open-set recognition algorithms. Best and the second best performing methods are highlighted in bold and italics fonts, respectively.}
	\label{table:osr}
\end{table*}

\begin{table*}[htp!!]
	\centering
	\begin{tabular}{|c|c|c|c|c|c|}
		\hline
		\begin{tabular}[c]{@{}c@{}}\textbf{Backbone Network}\end{tabular} & \textbf{Method} & \textbf{ImageNet-crop} & \textbf{ImageNet-resize} & \textbf{LSUN-crop} & \textbf{LSUN-resize} \\ \hline
		\multirow{8}{*}{\textbf{VGGNet}}
		& SoftMax             & 0.639          & 0.653          & 0.642          & 0.647          \\ %\cline{2-6} 
		& OpenMax             & 0.660          & 0.684          & 0.657          & 0.668          \\ %\cline{2-6} 
		& LadderNet + SoftMax & 0.640          & 0.646          & 0.644          & 0.647          \\ %\cline{2-6} 
		& LadderNet + OpenMax & 0.653          & 0.670          & 0.652          & 0.659          \\ %\cline{2-6} 
		& DHRNet + SoftMax    & 0.645          & 0.649          & 0.650          & 0.649          \\ %\cline{2-6} 
		& DHRNet + OpenMax    & 0.655          & 0.675          & 0.656          & 0.664          \\ %\cline{2-6} 
		& DHRNet + CROSR      & \textit{0.721} & \textit{0.735} & \textit{0.720} & \textit{0.749} \\ %\cline{2-6} 
		& MLOSR               & \textbf{0.837} & \textbf{0.826} & \textbf{0.783} & \textbf{0.801} \\ \hline
		\multirow{6}{*}{\textbf{DenseNet}}
		& SoftMax             & 0.693          & 0.685          & 0.697          & 0.722          \\ %\cline{2-6} 
		& OpenMax             & 0.696          & 0.688          & 0.700          & 0.726          \\ %\cline{2-6} 
		& DHRNet + SoftMax    & 0.691          & 0.726          & 0.688          & 0.700          \\ %\cline{2-6} 
		& DHRNet + OpenMax    & 0.729          & 0.760          & 0.712          & 0.728          \\ %\cline{2-6} 
		& DHRNet + CROSR      & \textit{0.733} & \textit{0.763} & \textit{0.714} & \textit{0.731} \\ %\cline{2-6} 
		& MLOSR               & \textbf{0.903} & \textbf{0.896} & \textbf{0.871} & \textbf{0.929} \\ \hline
	\end{tabular}
	 \caption{Open-set recognition on CIFAR10 for Out-Of-Distribution detection. Best and the second best performing methods are highlighted in bold and italics fonts, respectively.}
	\label{table:ood}
\end{table*}

The performance of the method is measured by its ability to identify unknown classes. Following the protocol from \cite{neal2018open}, the Area Under the ROC curve (AUROC) is used to measure the performance of different methods. The values reported in Table \ref{table:osr} are averaged over five randomized trials. The numbers corresponding to CROSR are taken from \cite{yoshihashi2018classification}. The results corresponding to all the other methods except MLOSR (proposed method), are taken from \cite{neal2018open}. CROSR \cite{yoshihashi2018classification} did not report its performance on the CIFAR10, CIFAR+10, and CIFAR+50 datasets, hence those numbers are not included here. The results on digits dataset are mostly saturated, and almost all methods perform more or less similar. CROSR achieves the best performance on digits dataset, with next best performance from MLOSR. However, for TinyImageNet which is much more challenging object classification dataset, MLOSR performs better than CROSR giving an improvement of $\sim$ 5\%.

\subsection{Experiment III : Out-Of-Distribution Detection}\label{subsec:ood_experiments}

In this experiment, we test the ability of MLOSR to identify OOD samples. Following the protocol defined by Yoshihashi \emph{et al.} in \cite{yoshihashi2018classification}, which uses in-distribution samples from CIFAR10 and samples from four different datasets (ImageNet-crop, Imagenet-resize, LSUN-crop and LSUN-resize) as OOD samples.  These four OOD datasets were developed specifically for CIFAR10 by \cite{liang2017enhancing}. Following the setup of \cite{yoshihashi2018classification}, OOD experiments use two backbone network architectures, VGGNet (referred in the paper as Plain CNN) and DenseNet. VGG, which consists of 13 layers, is a modified version of the VGG architecture as defined in \cite{simonyan2014very}. DenseNet follows the network architecture defined by \cite{huang2017densely} for CIFAR10. It has a depth of 92 and a growth rate of 24. Decoder architecture for all experiments is the same as used in the open-set experiments (modified to accommodate the image size), and the classifier architecture is a simple one layer fully-connected network with 10 neurons corresponding to 10 categories of CIFAR10 dataset, for all OOD experiments. We consider LadderNet \cite{valpola2015neural} and DHRNet baselines, where DHRNet architecture is a novel open-set method proposed by \cite{yoshihashi2018classification}. The performance is measured using F-measure (or F-1 score). All the images are of size $32 \times 32 \times 3$. For training we use all 50,000 training samples of CIFAR10 and evaluate the trained model on 10,000 OOD samples for each experiments, i.e., ImageNet-resize, LSUN-resize, ImageNet-crop, LSUN-crop. All the reported numbers except MLOSR are taken from \cite{yoshihashi2018classification}.

The Table \ref{table:ood} shows that for all OOD experiments MLOSR performs significantly better than other open-set algorithms. Second best performing method is DHRNet with CROSR both proposed in \cite{yoshihashi2018classification}. With VGGNet backbone, MLOSR is able to improve the performance by 11.6\%, 9.31\%, 6.3\% and 5.2\% for OOD samples from ImageNet-crop, ImageNet-resize,  LSUN-crop, LSUN-resize, respectively. On average, MLOSR with the VGGNet backbone performs better than the next best method by 8.05\%. Furthermore, MLOSR with DenseNet backbone significantly improves the performance by 17.0\%, 13.3\%, 15.7\% and 19.8\% for OOD samples from ImageNet-crop, ImageNet-resize, LSUN-crop and LSUN-resize, respectively. On average, MLOSR with DenseNet backbone performs better than the next best method by a significant 16.45\%. Overall MLOSR achieves 12.25\% improvement on average over the next best method (i.e., in both cases it is DHRNet+CROSR). Even though, there is no comparison provided for CIFAR10 dataset in Section \ref{subsec:openset_experiments} for open-set recognition experiments, results from OOD experiments show that for particular object datasets, MLOSR shows better performance than DHRNet+CROSR in identifying unknown samples.

\section{Conclusion}\label{sec:conclusion}
We presented an open-set recognition algorithm for deep neural networks called Multi-task Learning for Open-Set Recognition (MLOSR). MLOSR uses encoder, decoder and classifier networks trained in a multi-task framework.  We compare the performance of MLOSR with state of the art open-set algorithms and show better overall performance. Furthermore, we validate the ability of MLOSR to counter Out-Of-Distribution unknown samples by conducting experiments with the CIFAR10 dataset. MLOSR emerges as the best performing algorithm showing significant improvements over the baselines. Experiments show that MLOSR is able to deal with unknown samples better than recent competitive methods.  It achieves the state-of-the-art performance on various open-set recognition and OOD datasets.

\begin{acknowledgements}
This work was supported by the NSF grant 1801435.
\end{acknowledgements}

\bibliographystyle{spmpsci}      % mathematics and physical sciences
\bibliography{refs}   % name your BibTeX data base

\end{document}